\documentclass[preprint,12pt]{elsarticle}
\usepackage{amssymb}
\usepackage{geometry}
\newgeometry{left=2cm,right=2cm,top=2cm,bottom=2cm}
\usepackage{hyperref}
\hypersetup{colorlinks, citecolor=blue, linkcolor=blue, backref=true}
\usepackage{amsmath}
\usepackage{xcolor}

\usepackage{adjustbox}
\usepackage{caption}
\usepackage{subcaption}
\usepackage{algcompatible}
\usepackage{algorithm}

\journal{Expert Systems with Applications}
\begin{document}
\begin{frontmatter}
\title{Developing Hybrid Machine Learning Models to Assign Health Score to Railcar Fleets for Optimal Decision Making}
\author[mymainaddress]{Mahyar Ejlali\corref{mycorrespondingauthor}}
\cortext[mycorrespondingauthor]{Corresponding author}
\ead{mejlali@uchicago.edu}
\author[mysecondaryaddress]{Ebrahim Arian}
\author[mysecondaryaddress2]{Sajjad Taghiyeh}
\author[mysecondaryaddress3]{Kristina Chambers}
\author[mysecondaryaddress4]{Amir Hossein Sadeghi}
\author[mysecondaryaddress5]{Demet Cakdi}
\author[mysecondaryaddress4]{Robert B Handfield}

\address[mymainaddress]{The University of Chicago, Chicago, IL USA \\ email: \href{mejlali@uchicago.edu}{mejlali@uchicago.edu} }
\address[mysecondaryaddress]{University of Illinois at Urbana-Champaign, Champaign, IL USA \\ email: \href{mailto:Arian2@illinois.edu}{Arian2@illinois.edu}}
\address[mysecondaryaddress2]{North Carolina State University, Raleigh, NC USA \\ email: \href{mailto:Staghiy@ncsu.edu}{Staghiy@ncsu.edu}}
\address[mysecondaryaddress3]{TTX company, Chicago, IL USA \\ email: \href{mailto:Kristina.Chambers@ttx.com}{Kristina.Chambers@ttx.com} }
\address[mysecondaryaddress4]{North Carolina State University, Raleigh, NC USA \\ email: \href{mailto:Asadegh3@ncsu.edu}{Asadegh3@ncsu.edu}}
\address[mysecondaryaddress5]{TTX company, Chicago, IL USA \\ email: \href{mailto:Demet.Cakdi@ttx.com}{Demet.Cakdi@ttx.com} }
\address[mysecondaryaddress4]{North Carolina State University, Raleigh, NC USA \\ email: \href{mailto:Rbhandfi@ncsu.edu}{Rbhandfi@ncsu.edu}}

\begin{abstract}
A large amount of data is generated during the operation of a railcar fleet, which can easily lead to dimensional disaster and reduce the resiliency of the railcar network. To solve these issues and offer predictive maintenance, this research introduces a hybrid fault diagnosis expert system method that combines density-based spatial clustering of applications with noise (DBSCAN) and principal component analysis (PCA). Firstly, the DBSCAN method is used to cluster categorical data that are similar to one another within the same group. Secondly, PCA algorithm is applied to reduce the dimensionality of the data and eliminate redundancy in order to improve the accuracy of fault diagnosis. Finally, we explain the engineered features and evaluate the selected models by using the Gain Chart and Area Under Curve (AUC) metrics. We use the hybrid expert system model to enhance maintenance planning decisions by assigning a health score to the railcar system of the North American Railcar Owner (NARO). According to the experimental results, our expert model can detect 96.4\% of failures within 50\% of the sample. This suggests that our method is effective at diagnosing failures in railcars fleet. 
\end{abstract}

\begin{highlights}
\item An expert hybrid predictive fault method is proposed based on fast-DBSCAN and PCA.
\item Enhancing maintenance planning decisions by assigning a health score.
\item Inspection data from 1986–2020 of North American Railcar Owner (NARO) is used.
\item The model is able to predict future faults in the railcar fleet accurately.
\end{highlights}

\begin{keyword}
Expert system, Predictive maintenance, Railcar maintenance, Machine learning, Maintenance health score
\end{keyword}

\end{frontmatter}

\section{Introduction}\label{intro}
Maintenance consists of activities that ensure the railcar assets continue to operate safely and reliably. These activities include inspection, repair, testing, and replacement of parts. Maintenance actions seek to increase operational availability by reducing instances in which the asset is unavailable for use. An effective and efficient maintenance strategy is vital because there is often a high cost associated with out-of-service maintenance tasks \citep{mohammadi2022deep}.

Predicting the requirements for maintaining equipment helps practitioners develop an efficient plan to perform maintenance before a mechanical deterioration occurs and avoid unnecessary out-of-service time \citep{wan2017manufacturing}. Most systems contain mechanical building blocks subject to wear and tear, eventually becoming faulty if not appropriately maintained. The deterioration process of components does not occur at a uniform rate and depends on the system's conditions, such as environment, stress, and load. The railway industry has a growth rate of +1.5\%, making it the second most rapidly growing transportation module \citep{bukhsh2019predictive}. The need for improved planning for the maintenance of railcar assets is increasing, given that the components of railcars deteriorate more quickly as usage and demand grow \citep{de2016railway}.

Ensuring safe and reliable assets while achieving optimal cost is a primary objective for maintenance planning. Corrective maintenance and regularly scheduled preventative maintenance comprise the traditional maintenance approaches \citep{marquez2007maintenance}. Corrective maintenance strategies incur higher costs and longer operational delays, while the repetitive maintenance approach results in additional costs as maintenance happens before needed. Another widely adopted strategy is condition-based maintenance. In this approach, the asset's current state is considered part of maintenance decisions \citep{al2016improvement}. These decisions are made based on quantitative engineering analysis, a repetitive schedule, and budgetary considerations.

There is a significant cost associated with an unplanned maintenance event. If component degradation could be predicted ahead of time, it would allow sufficient time for operations to inspect a given problem and fix it before it causes an out-of-service maintenance event. Many alert systems have been developed in the United States to manage the prevention and prediction of failures \citep{li2015prediction} A significant interruption to existing rail operations could occur if worn or defective components are not identified and prevented \citep{shafiullah2010predicting}.

Data analytic is becoming an important technology within the railway sector \citep{cirovic2013decision,ghofrani2018recent,oneto2020dynamic,spigolon2019improving,oneto2019restoration}. With recent advancements in technology, it has become feasible to store and manage a large amount of data to manage, operate and maintain transportation infrastructures through business intelligence \citep{mazzarello2007traffic,turner2016review,an2023hybrid}.

In \cite{pipe2016automated} and \cite{lee2016universal}, the authors evaluated the current and future status of assets by applying data-driven methods to predict functionalities. In other research studies \citep{liden2015railway,consilvio2015modular,faris2018distributed}, decision support algorithms were embedded into railway asset management practices to minimize maintenance costs while planning maintenance strategies  \citep{d2019integrated}. However, the potential for embedding data-driven methods into a predictive maintenance decision support system to calculate health rates for railcar components seems to have not yet been fully explored.

Maintenance approaches can be categorized into three main categories: 1. Corrective maintenance (run-to-failure); 2. Preventive maintenance; and 3. Predictive maintenance (PdM). The first approach, corrective maintenance, is the simplest and least desirable method. In this approach, the component is replaced once it is no longer viable and cannot function. Corrective maintenance can have drawbacks, such as unplanned downtime, missed customer commitments, and higher total ownership cost. In the second approach, preventive maintenance, maintenance activities are performed according to planned periodic time intervals. These intervals are usually defined according to engineering analysis and visual inspection, and the replacement occurs without considering the component's health status \citep{malik1979reliable}. This strategy may be expensive, as complex systems of components have varying life cycles based on their use condition. 

Moreover, the preventive maintenance strategy does not provide information on the component of health status, useful data for planning, and cost optimization. Predictive maintenance considers critical components to predict their “probability of failure.” The probability of failure designates that the part is worn beyond the industry-recommended usage limits. Based on these predictions, appropriate measures are proactively applied to address any maintenance issues before they occur. This method's advantage is to minimize maintenance costs while extending the useful life of components \citep{qiao2015survey}. Unlike corrective maintenance, predictive maintenance achieves significant cost efficiencies by avoiding unplanned downtime, unnecessary maintenance, and unexpected parts replacement \citep{bukhsh2018machine}.

\begin{figure}[h]   
\centering
  \includegraphics[width=15cm]{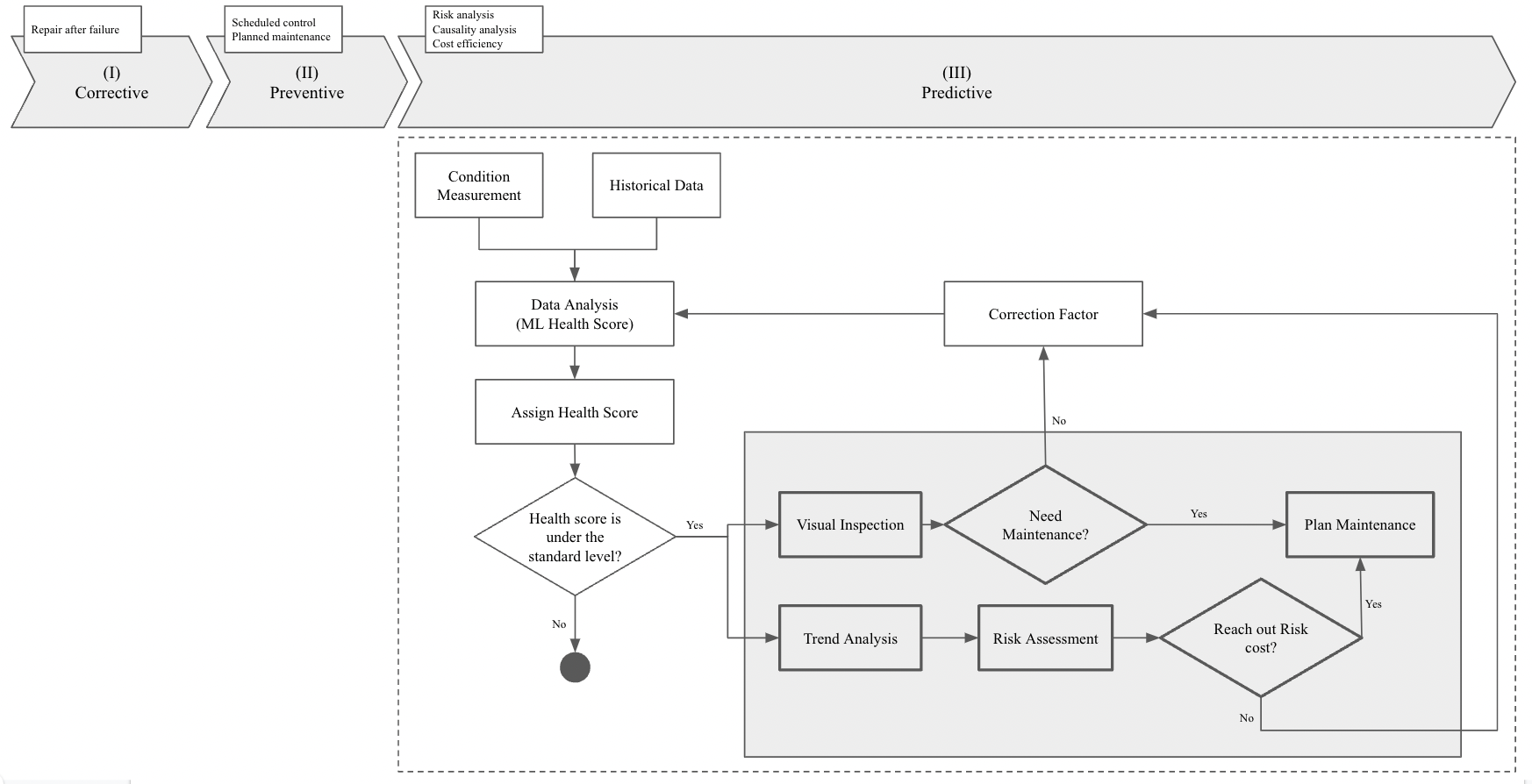}\\
  \caption{Proposed framework for predictive maintenance in railcar fleets.}
  \label{fig:process flowchart}
\end{figure}

Machine learning (ML) methods have shown promising applications across various industries by developing models that rely on large datasets \citep{bose2001business,yekkehkhany2019risk,taghiyeh2020forecasting,taghiyeh2020multi,taghiyeh2021loss,obermeyer2016predicting}. In supervised ML methods, historical data can predict future states based on detecting patterns and learning from this data. For example, inspection and condition assessment historical data is utilized for training an ML classifier to indicate components' future conditions to classify them as well-maintained or at the end of their useful life. Additionally, ML methods can reveal underlying characteristics of the distribution of historical data, which is used for many applications such as outlier detection and time series forecasting. These models can help decision-makers predict maintenance needs in advance and avoid unnecessary or unplanned downtime. Figure~\ref{fig:process flowchart} shows the proposed model in this paper for the predictive maintenance framework.

There is a need in the railroad industry for decision-support models to help practitioners manage their maintenance strategies more efficiently, based on available data from diverse sources. To improve the process of maintenance decision-making while reducing uncertainty, we propose leveraging machine learning techniques to build a predictive maintenance framework. This approach requires a large amount of historical data on visual inspection records, asset specifications, and maintenance records. The model's output provides a health condition metric (rate) associated with each railcar, helping practitioners make proactive maintenance decisions. The remainder of this paper is organized as follows. In Section~\ref{sec:litRev}, we review the literature on maintenance strategies. Section~\ref{sec:Proposed Framework} shows the machine learning methodology applied to develop the health rating for each railcar. Section~\ref{sec:Numerical Study} evaluates the proposed health score model performance, using real data provided by NARO. Finally, conclusions are discussed in Section~\ref{sec:Conclusion}.

\section{Literature Review}\label{sec:litRev}
Maintenance policies have been the subject of much academic and industry research \citep{mccall1965maintenance,wang2002survey,shin2015condition,iqbal2017inspection}. Several categorizations for maintenance management policies can be found in the literature. In this paper, we focus on the categorization proposed by \cite{mccall1965maintenance,wang2002survey,shin2015condition,iqbal2017inspection}. Hence, we can classify the maintenance strategies into the following categories:

\begin{itemize}

    \item Run-to-Failure (R2F) or correction maintenance: This strategy's underlying logic is to perform maintenance when a component stops working, making it the most uncomplicated approach among all policies. This policy can be the most expensive, as it may cause sudden disruptions in workflow.  
    
    \item Preventive Maintenance (PvM): This strategy is also known as scheduled maintenance or time-based maintenance, in which the component is scheduled for maintenance according to a time or iteration-based period. It can effectively avoid failures, but the downside is that it may lead to an additional operating cost since unnecessary corrective actions may need to be taken.   
    
    \item Predictive Maintenance (PdM): In this approach, predictive tools are utilized to forecast necessary maintenance actions by continuously monitoring components' performance and performing maintenance as needed. Additionally, using statistical tools such as machine learning techniques based on historical data and engineering approaches, this strategy allows  hi for the early detection of failures.
    
\end{itemize}

When industries decide to use R2F, they assume the risk of their assets' unavailability by delaying maintenance actions. Moreover, while using the PvM strategy, components are replaced before their useful life. A good maintenance strategy needs to minimize maintenance costs while maximizing the component's life.  Hence, the PdM is a strategy that shows the most promising applicability among all other approaches \citep{jezzini2013effects}, thus attracting the most attention in recent years \citep{kumar2019hmm}. PdM's key advantages include reducing maintenance activities, maximizing the component's operation time, and minimizing labor and material costs.

Recent advancements in the industry, such as the prevalence of condition monitoring equipment and the availability of deterioration prognosis and residual life estimation approaches, have made predictive maintenance possible \citep{bousdekis2018review}. In theory, PdM is similar to condition-based maintenance (CBM); however, in PdM, online health prognostic information is used as the basis for maintenance decisions, while in CBM, online diagnostic information is used \citep{ahmad2012review}. PdM's goal is to efficiently predict system failures to intervene and perform necessary maintenance actions promptly. Various PdM methods have been proposed in the literature \citep{van2009survey,wang2011overview,wang2012overview,ahmad2012overview,sanchez2016maintenance}.

Research studies on PdM evolved from visual inspection methods to automatic and advanced techniques, such as machine learning, pattern recognition, and neural networks. These advanced PdM methods have proven to be useful in many industries, including the railway industry, by detecting and collecting equipment's sensitive information that could not be tracked manually by a technician \citep{hashemian2010state}. PdM reduces downtime using integrated sensors, avoiding unnecessary maintenance, improving efficiency, and saving cost. In some ways, PdM and PvM overlap, as both plan maintenance actions in advance to avoid abrupt failures. However, PdM maintenance schedules are based on a prognosis derived from collected data and relevant algorithms \citep{wu2007neural,frontoni2017hdomo}.

\cite{jardine2006review} classified three categories of diagnostic and prognostic models used in maintenance strategies: statistically-based models, artificial intelligence-based models, and model-based methods. From these three categories, artificial intelligence models are increasingly being applied in PdM applications. Statistics-based models require a strong mathematical background, and mechanistic knowledge is needed to use model-based methods. For instance, in a study performed by \cite{baptista2018forecasting}, everal artificial intelligence approaches were compared to the life usage model, a statistical technique that uses the same concept as PdM. The results demonstrated that artificial intelligence methods display better performance when compared to statistical methods. The artificial intelligence and machine learning (ML) models are data-driven approaches and use historical data to train the current model and predict the component's future state \citep{nguyen2017model,wang2016wind,medjaher2012remaining,chu2015incrementaal}. In general, these data-driven models require a large amount of historical data to train the underlying models \citep{paolanti2017visual,paolanti2015automatic,paolanti2017mobile,naspetti2016automatic,yekkehkhany2021cost}.

We can categorize ML-based PdM methods into two major classes: supervised and unsupervised learning. In supervised learning algorithms, failure information is available in the dataset for modeling. In unsupervised learning methods, only the logistical information is available, and there is no information on maintenance actions or failures \citep{susto2014machine,wang2016intelligent}. Whether or not maintenance information is available usually depends on management policy concerning data collection and availability. However, supervised learning approaches are preferable.
Supervised learning methods can also be divided into two categories: regression and classification. If the output is continuous, regression models are used, but we use classification models to deal with categorical output values \citep{druck2008learning}.

Several ML-based research articles exist in PdM literature focused on railway track performance or one specific railcar component. In \cite{de2016railway}, recurrent neural networks were used to track and find failures in railway track circuits by incorporating data from their signals. Heterogeneous data was used in \cite{li2014improving} to develop failure prediction models and improve railway network velocity. These heterogeneous data include maintenance, inspection, and failure data. In a similar study, diagnostic log data were used by \cite{kauschke2014learning} to predict a train's component failure. Additionally, studies have used diverse data regarding relevant events to improve historical information quality \citep{nunez2014facilitating}. ML models can help decision-makers use a large amount of historical information from various sources and embed it in a learning model to predict assets' future state.

There are relatively few articles in the literature that focus on using ML techniques for PdM's maintenance approach applied to an individual railcar. In \cite{manco2017fault}, an ML solution for detecting the failures of a metro train door was investigated. \cite{bohm2017remaining} proposed a method to forecast the remaining useful time of a switch based on the switch engine's power consumption. \cite{chen2017automatic} used a convolutional neural network model to detect defects in the fasteners supported on the catenary device. These studies show an increasing trend toward using ML techniques for PdM in the railway industry. However, the presented methods are tailored for a specific component and cannot be extended for other maintenance applications. Additionally, most of these studies take extra data collection steps, such as condition monitoring devices, which are expensive and impractical \cite{de2016railway}. Moreover, sophisticated ML methods such as neural networks and deep learning models are challenging to interpret for infrastructure managers, although they provide accurate results.

This study uses machine learning techniques, specifically classification methods such as random forest and decision trees, to develop an expert system to assign health scores to rail cars. We identify critical components within a system and then determine the probability of failure for each. The second part of the process is assigning a criticality factor to each component to calculate a composite health rating. The developed health score helps practitioners in their decision-making process to prioritize railcars for maintenance.

\section{Proposed Framework} \label{sec:Proposed Framework}
Companies in the rail industry are leaning towards using predictive maintenance over preventive maintenance frameworks due to its cost efficiency.  Our framework develops a machine learning-based predictive maintenance application to assign a health rating, representing the relative need for maintenance for that asset group. We use the health rating of railcars as an indicator of future maintenance needs. The health rating could also be applied as a decision-making tool for fleet sizing, including optimization. In this section, we first provide a background for each selected component of a railcar to show the impact of the respective components' failure. Next, we introduce the general framework and describe each feature of the model in detail.

\subsection{Background}
A railcar health rating provides a single source of information from which management can identify trends to support long-term fleet reliability and make maintenance decisions. The health score uses real-time railcar data to provide timely identification of maintenance needs related to critical components. For this study, we focus on a specific railcar type and four components suggested by experts from a NARO in the railroad industry. We apply machine learning techniques to predict the probability of required maintenance and assign health rates to individual railcars.

\subsection{General Framework}
\begin{figure}[h]   
\centering
  \includegraphics[width=15cm]{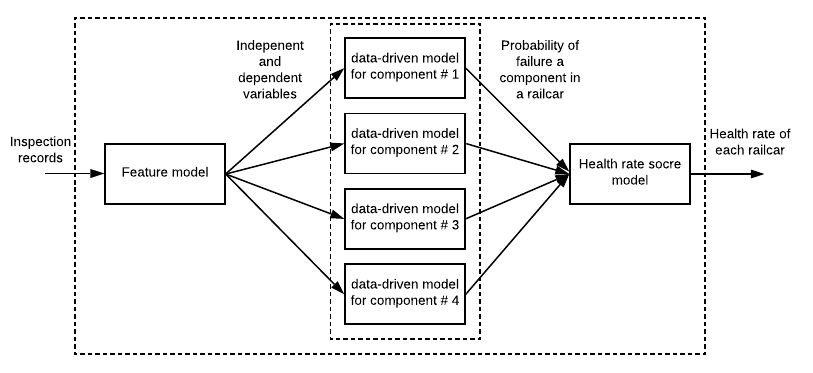}\\
  \caption{Proposed framework for feature selection.}
  \label{fig:Proposed Framework}
\end{figure}

The proposed predictive maintenance framework aims to find the railcars that are most likely to require significant maintenance. This goal helps managers to allocate their resources efficiently in identifying railcars that need attention. The framework has three critical models, as shown in Figure \eqref{fig:Proposed Framework}. 

In the feature model, we use a cut-off time to define the independent and dependent variables, based on each component's historical failure data. We also engineer some independent features to obtain more information, which is explained in detail. We develop four different classification models for each component to detect the failure and non-failure observations in the data-driven model. Using the outputs of these models we identify the failure probability of a component in the railcars. We also employ two different metrics, Area Under Curve (AUC) and Gain Chart, to evaluate the model for each component. The logic for selecting the metrics are explained in details. If the results from the evaluation metrics are satisfactory, we move to the third model. In the health score model, we use the output for each component's classification model to calculate the health score corresponding to each railcar, helping the decision-maker to select the correct railcar for maintenance.

\subsection{Feature Model}
\begin{figure}[h]   
\centering
  \includegraphics[width=15cm]{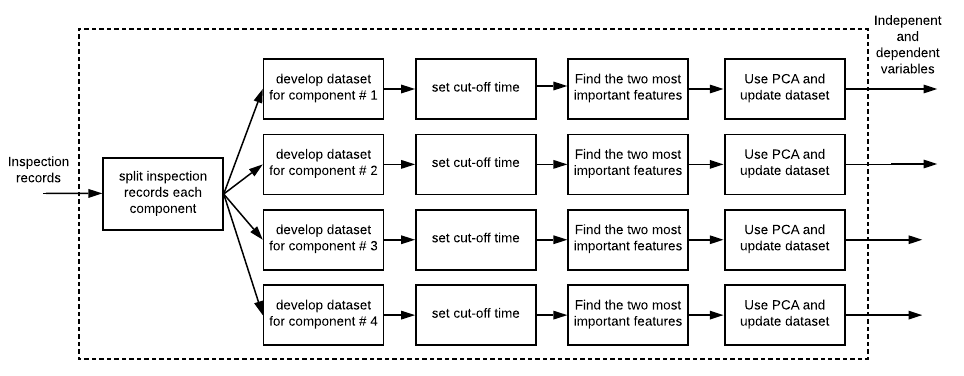}\\
  \caption{Proposed Feature model.}
  \label{fig:feature_model}
\end{figure}

A comprehensive maintenance dataset associated with four railcars components were acquired from a North American railcar owner (NARO) and used in this study.  Each record in the dataset represents several railcar and component features. Some examples are railcar age, geographical loading history, miles traveled, components replacement history, and component condition (new or refurbished). In predictive maintenance, we need to use historical data to predict whether or not a component in a railcar is going to fail in the upcoming years. To achieve this goal, we first split the components' data, as shown in Figure \eqref{fig:feature_model}. Thus, each component has a dataset that corresponds to its own maintenance history and characteristics.

\begin{figure}[h]   
\centering
  \includegraphics[width=15cm]{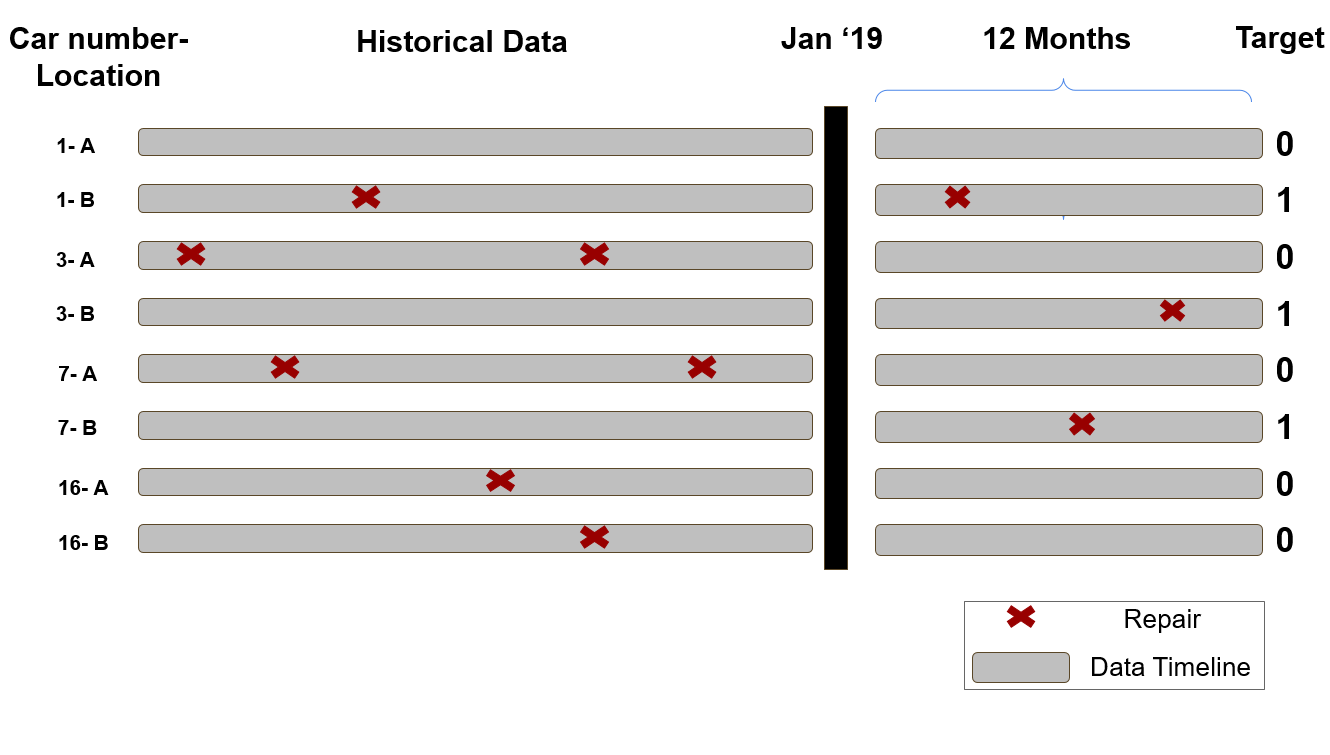}\\
  \caption{Dependent variable (target) setup procedure.}
  \label{fig:in_dep_feature}
\end{figure}

We then set the cut-off time to verify that each component's model can adequately predict future failures. As shown in Figure  \eqref{fig:in_dep_feature}, January 2019 is the cut-off date. We also join the railcar number with the railcar component location to use as an ID associated to each sample.

All of the components in this study have two locations, whereby the corresponding inspections are performed separately on each location. The dependent variable (target) for each component dataset is whether or not a railcar component has been identified as failing after January 2019. Let $J$ and $I$ be the number of railcars, and components, respectively. We denote $y_{ij}$ as the dependent variable such that

\begin{equation*} 
    y_{ij} = \begin{cases} 
    1 & \text{if the component i of railcar j failed after the cut-off time}, \\
    0 & \text{Otherwise}.
    \end{cases}
\end{equation*}

Several new independent features for each railcar have been defined, based on data before the cut-off date. We introduce nearly sixty different independent features, such as the average number of failure records, component ages, and mileage since the last repair. Some of these features are shown in Table~\ref{tab:feature model}.

\begin{table}[h]
\centering
\caption{Dependent features of the model.} 
\label{tab:feature model}
\begin{tabular}{cc}
\hline
 Component features & Railcar features \\
\hline
Mileage on component since last replacement & Car age \\
Component age & Loading count \\
Pocket number & Loading regions \\
Condition code  & Average days in service \\
\hline
\end{tabular}
\end{table}

\subsubsection{Clustering based on fast-DBSCAN}
We apply pre-processing techniques to impute missing data and use the fast DBSCAN algorithm introduced by \cite{hanafi2022fast} method to cluster categorical data that are similar to one another within the same group.

DBSCAN \citep{ester1996density} is a density-based clustering algorithm that works by identifying points in the dataset that are in high-density regions, and using them to expand clusters. It starts by selecting a point at random and checking whether it has enough neighbors within a specified distance ($\epsilon$) to be considered a "core point." If it does, it is added to a cluster and the algorithm looks for other points in the dataset that are within $\epsilon$ distance of this point and also have at least $Min_{Points}$ neighbors. These points are also added to the cluster, and the process is repeated until all reachable points in the cluster have been found. If a point does not have enough neighbors within $\epsilon$ distance to be considered a core point, it is labeled as noise. DBSCAN is a very effective algorithm for clustering and imputing missing data in a dataset. However, in the case of big data, the DBSCAN algorithm needs heavy computations, which slows down clustering and increases run time \citep{hanafi2022fast}.

The proposed algorithm by \cite{hanafi2022fast} starts with introducing the concept of the operational dataset (called OP), and potential dataset (named PD). The operational dataset includes all data within a sphere of size radius centered at point P and PD excludes the data existing in the operational dataset. Equations~\eqref{eq:op set}, \eqref{eq:pd set} represent the mathematical formula for OP and PD respectively.

\begin{flalign}
    \label{eq:op set}  & OP = \{ x | x \in D, |x-P| \leq size \}, n \in R, size = n \times \epsilon, n \geq 1 \\ 
    \label{eq:pd set}  & PD = D - OP &&
\end{flalign}

In this context, $D$ represents the initial database, $x$ is a sample within the database, $P$ is the center of the OP, and $|x-P|$ represents the Euclidean distance between data $P$ and $x$. The radius of the operational dataset is represented by $n \times \epsilon$.

Here are the three steps involved in performing fast-DBSCAN: 

\begin{enumerate}

    \item  In the initial step of this procedure, the primary dataset is split into two sections. One point, designated as $P$, is randomly selected from the primary dataset and the distance from P to every other point in the dataset is computed. Based on the values of the parameters $\epsilon$ and $Min_{Points}$, $P$ may be classified as noise or a core point. $\epsilon$ parameter refers to the radius of the DBSCAN algorithm and minimum number of points ($Min_{Points}$) is a parameter in the DBSCAN algorithm that is used in conjunction with the radius ($\epsilon$) to determine whether a point is a noise or a core point. If a point has at least $Min_{Points}$ number of other points within the $\epsilon$ radius, it is considered a core point. If it has fewer than $Min_{Points}$ within the radius, it is considered noise. This parameter helps in identifying the density of points in a cluster and removes noise points from the dataset. If $P$ is identified as a core point, all points within a certain distance, as determined by the user-specified radius of the $OP$, are considered part of the $OP$, while the remaining points are part of the $PD$. The distance between points in the OP is then calculated to identify which points are neighbors of $P$. This approach reduces computational cost when compared to the traditional DBSCAN method.
    
    \item Once the OP is formed, the DBSCAN algorithm is applied to it. The process involves going through each point in the OP and identifying its neighboring points. If a point is found to be a core point, all its neighboring points are added to a list. A point is then removed from the list, and the procedure is repeated until no more points remain in the list. This leads to the formation of a list of data points that are considered to be part of the same cluster.
    
    \item In the final stage of this procedure, the OP is updated. This is required as the process of determining the distance from other points and identifying neighbors is restricted to the OP that a specific point belongs to. In order to scan and locate the neighbors of data points that are not currently included in the OP, the OP must be updated.
    
\end{enumerate}

After using the above algorithm for imputing missing data, we use PCA for feature extraction, and finally provide an algorithm to calculate the health score based on the component and railcar features. 

\subsubsection{Feature extraction based on PCA} \label{sec:fastDBSCAN}
Principal component analysis (PCA) is a statistical technique that is used to reduce the dimensionality of a data set by projecting it onto a lower-dimensional space while retaining as much of the original information as possible \citep{ali2021automatic, ma2018automatic}. PCA is often used in predictive maintenance and calculating health scores because it can help identify patterns and trends in the data that may not be immediately apparent using other techniques. It can also help reduce the complexity of the data and make it easier to interpret \citep{calvo1998comparative}.

In PCA, the data set is represented in a wide matrix form, with rows representing observations and columns representing variables. The PCA transformation converts this data set into a set of PCs, which are essentially a set of coefficients that can be interpreted as the projections of the observations onto the principal components (PCs). These PCs are ordered in terms of their importance, with the first PC being the most important and the last PC being the least important. 

Let $\textbf{x}_{ij} = [x_{ij1},...,x_{ijK}]$  be the dependent features vector of component $i$ of railcar $j$, where $K$ is the number of dependent features. We categorize these dependent features based on the component and railcar features. 

Here are the four steps involved in performing PCA on a data set \citep{jolliffe2002principal}:

\begin{enumerate}

    \item Standardize the data: The first step is to standardize the data, which means to center and scale it so that each variable has a mean of zero and a standard deviation of one. This is important because PCA is sensitive to the scale of the variables, and standardizing the data helps ensure that all variables are on the same scale. Arithmetic mean and standardized matrix are calculated in \eqref{eq:mean}, \eqref{eq:standard} respectively.
        \begin{flalign}
            \label{eq:mean}  & \Bar{x}_j = \frac{1}{n} \sum_{i=1}^{n} x_{ij}\\ 
            \label{eq:standard}  & B_{ij}=x_{ij}-\Bar{x}_j&&
        \end{flalign}
        
    \item Calculate the covariance matrix: The next step is to calculate the covariance matrix of the standardized data \eqref{eq:cov matrix}. The covariance matrix is a measure of the relationship between pairs of variables, and it helps us understand how the variables are related to each other.
        \begin{flalign}\label{eq:cov matrix}
            S_B = \frac{1}{n-1} B^T B &&
        \end{flalign}
        
    \item Calculate the eigenvalues and eigenvectors of the covariance matrix: The eigenvalues ($\lambda_j$) and eigenvectors ($p_j$) of the covariance matrix are used to determine the principal components (PCs) using equation~\eqref{eq:eigen}. The eigenvalues represent the amount of variation explained by each PC, and the eigenvectors represent the directions along which the maximum variation occurs.
        \begin{flalign}\label{eq:eigen}
            S_B p_j= \lambda_j p_j &&
        \end{flalign}
        
    \item Select the number of PCs to retain and transform data: Select the number of PCs to retain (highest to lowest), then we can use the PCs to transform the original data into the lower-dimensional space defined by the PCs.
        
\end{enumerate}

Let $x^{PCA}_{ij} = [x^{PCA}_{ij1},...,x^{PCA}_{ijm}]$ be the PCA features vector of component $i$ of railcar $j$. Then, new vector of independent feature vector of components $i$ of railcar $j$ is $\Tilde{\textbf{x}}_{ij} = [x_{ij1},...,x_{ijK},x^{PCA}_{ij}]$. In the next step of the feature model, we explore which new independent features have more information to classify non-failure and failure events. We use a data-driven model to find the most important features in the dataset. 

\subsection{Data-Driven Model}
The data-driven model is a mathematical function that estimates the actual function of interest as accurately as possible.  Let $\hat{f}_i$ be the data-driven model of component $i =1,..,I$, such that $\Tilde{\mathbf{x}}_i = [\Tilde{\mathbf{x}}_{i1}, ..., \Tilde{\mathbf{x}}_{iJ}]$ is independent feature variables matrix , and $\mathbf{\hat{y}}_i$ is the predicted outcome of the component $i$ model.

\begin{flalign}
    \label{eq:data-driven-model} \hat{f}_i: \Tilde{\mathbf{x}}_i \rightarrow \mathbf{\hat{y}}_i \quad  \forall i = 1,..,I
\end{flalign}

For each component i, we build a classification model to predict whether a railcar component will fail after the designated cut-off time. The model's input includes the information about the railcar before the cut-off date $\Tilde{x}_i$, and the output is the probability of the failure of component $i$ in car number $j$, $p_{ij}$. In the training process, we use a set of railcars randomly selected to train the model. To test the model, we use the remaining railcars to validate our data-driven model for the components.

\subsection{Evaluation Metrics }
To evaluate and enhance the effectiveness of the developed models, we apply AUC and Gain Chart metrics. Gain Chart is a visual presentation to measure the results obtained with or without the predictive model. Specifically, this metric allows NARO to manage budget restrictions. When the company is limited to inspecting a portion of the fleet, the Gain Chart represents the number of failures captured within the selected sample compared to random selection. The other method we use is AUC, which provides an aggregate measure of performance across all possible classification thresholds. When AUC is higher, the model distinguishes non-failures and failures with a higher degree of probability.

False-positives show the number of actual failures, which are predicted as non-failure records. False-negatives demonstrate the number of actual non-failures, which are predicted as failure records. Our dataset is highly imbalanced, and the failure rate is less than 1\% for some of the components. Within this imbalanced dataset, the false-negative rate is significantly higher than the false-positive rates. We use AUC to find the optimal threshold by evaluating the true-positive rate against the false-positive rate. We set a threshold as a decision variable to balance false-positive and false-negative values for a given hyper-parameter of the data-driven model. Let $t$ be the threshold. The objective is to maximize the AUC of a given data-driven model $f(.)$. 

\begin{flalign}
    \label{eq:auc} AUC(f) = \max_t AUC(t,f) 
\end{flalign}

We denote  $AUC(t,f)$ as the AUC function of the data-driven model $f$ with threshold $t$.. We use the above-listed optimization problem to find the maximum AUC for a given $f$, defined as $AUC(f)$.

\subsection{Heath Rate Model}
In the previous steps, we independently created predictive models for each component. The probabilities for components' failure are aggregated to assign a health score to railcars. Let $p_{ij}$ be the
probability of the failure of component $i$ in railcar number $j$, which is the output of the data-driven model of component $i$. We denote $w_i$ as the component critically associated with component $i$. Therefore, the health score (HR) for each car number $j$ is defined as follows ($i$ corresponds to the components):

\begin{flalign}
    \label{eq:health-rate} HR_{j} =  \sum_{i}w_ip_{ij} 
\end{flalign}

\begin{figure}[h]   
\centering
  \includegraphics[width=15cm]{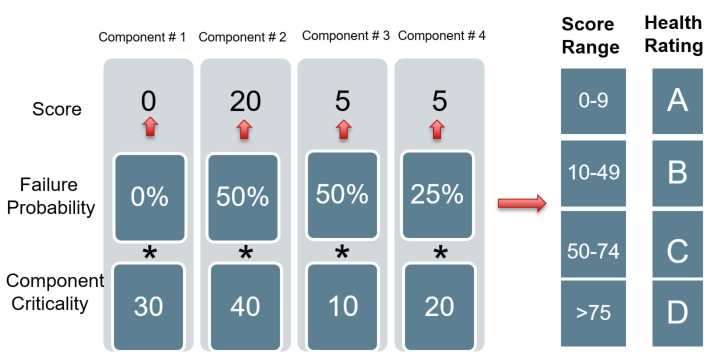}\\
  \caption{Health score model.}
  \label{fig:health}
\end{figure}

The health score calculation process is shown in Figure  \eqref{fig:health}.

Expert railcar engineers determine the importance of components. The weights are used to generate the health score. Algorithm~\ref{alg:framework} shows the predictive maintenance framework used to assign a health score to each railcar.

\begin{algorithm}[hbt!]
\caption{Proposed predictive maintenance framework}\label{alg:framework}
\begin{algorithmic}

\STATE \textbf{Step 1.} Imputation of the missing values using fast-DBSCAN \citep{hanafi2022fast} 
\STATE \textbf{Input:} $\epsilon, Min_{Points}, P, N, C$
\STATE \textbf{Output:} Clustered groups, Impute missing value

\WHILE{$N \neq \emptyset $}
\STATE $V \gets$ Top of Neighbor

\IF{($V$ is not visited) $\land$ ($V\in OP$)}
    \STATE Visit V;
    \STATE $V_{OP} = V_{OP} + 1$;
    \STATE $NB \gets$ Distance of V and Sample from OP;
    \STATE Find the $\epsilon$-neighborhood of V;
\ELSIF{$|NB| \geq Min_{Points}$}
    \STATE $N \gets N + |NB|$
\ENDIF
\STATE $Candid \gets Candid + V$
\ENDWHILE
\STATE Add V to the cluster of C 

\WHILE{$V_{OP} \leq M - V_B$}
\STATE Expand cluster using method introduced in section~\ref{sec:fastDBSCAN}
\STATE Assign cluster value to the missing feature value
\ENDWHILE

\STATE \textbf{Step 2.} Set the cut-off time to define independent variables matrix $\textbf{x}_i$ and the target vector $\textbf{y}_i$. 
\STATE \textbf{Step 3.} Find the important feature and use PCA to add a new independent variable. Update independent variables matrix $\tilde{x_i} = (x_i, x^{PCA}_i)$. 
\STATE \textbf{Step 4.} Separate the dataset of each component to train and test sets.

\STATE $i \gets 1$
\WHILE{$i \neq I$}
\STATE Train each data-driven model $f_i$
\STATE Calculate $p_{ij}$
\STATE $i \gets i+1$
\ENDWHILE

\STATE \textbf{Step 5.} Calculate the health score of each car number $j$ which is $HR_{j} =  \sum_{i}w_ip_{ij}$.

\end{algorithmic}
\end{algorithm}

\section{Numerical Study} \label{sec:Numerical Study}
In this section, we use the data provided by NARO to validate the proposed predictive maintenance framework. We review the dataset in detail and analyze the features of the data. We explain the engineered features and evaluate the selected models by using the Gain Chart and AUC metrics. Ultimately, we use the best model to assign health rates to railcars.

\subsection{NARO Dataset}
The variables extracted from the dataset include the maintenance history, car properties, and movement history of railcars. The repair and inspection data are collected for four components of a specific railcar type from 1986 to 2020. The selected fleet for our study contains approximately 11,000 active railcars. We apply pre-processing techniques to impute missing data and use the DBSCAN method to cluster categorical data that are similar to each other within the same group.

\subsection{Feature model}
As we discussed in Section 3, to develop the maintenance framework, we needed to set up a dataset for each component from the dataset to predict and identify potential maintenance needs in the following year, based on historical data. Therefore, the cut-off time was set to January 2019, and we predicted required maintenance for the upcoming year. The next step was to generate the dependent feature in the dataset, which is failure or non-failure. We defined this binary feature according to occurrence of a failure after the cut-off date. If a railcar component failed after January 2019, the dependent feature of the railcar would be set to 1, which is associated with a failure; otherwise, it would be set to 0, which shows a non-failure state for the corresponding component.

Additionally, based on the dataset before the cut-off date, several independent features were developed. These features can be categorized as railcar features and component features. Some of these features are shown in Table~\ref{tab:feature model}. Random forest classifier is used to identify the important features in the datasets corresponding to each component, and the results are shown in Table~\ref{tab:feature-importance}.

\begin{table}[h]
\centering
\caption{Important features by components.} 
\label{tab:feature-importance}
\begin{adjustbox}{width=\textwidth}
\begin{tabular}{cccc}
\hline
Component \#1 & Component \#2 &  Component \#3 & Component \#4 \\
\hline
component age & component age & component age & mileage since last replacement \\
car age &  average day of trip being loaded & mileage since last replacement & average day of trip being empty\\
mileage since last replacement & car age & car age & car mileage\\
car mileage & mileage since last replacement & average day of trip being empty & component age\\
average day of trip being loaded & car mileage & car mileage & car age\\
\hline
\end{tabular}
\end{adjustbox}
\end{table}

It is interesting to note that "component age," "car age," "mileage since last replacement," and "car mileage" are the four most important features among the top five found in all datasets. Also, "component age" and "mileage since last replacement" show correlations nearly equal to 0.95 in all datasets. "Car age" and" car mileage" have a correlation of 0.87 in all datasets. Therefore, we build our random forest classifier model without correlated counterparts. We observed "mileage since the last failure" is one of the most critical features in the model, which results in leaving out the" component age" factor. Another model that does not include "mileage since the last failure, finds "component age" is one of the most important features. This is also true for "car age" and "car mileage." Hence, we use the PCA model to create two features based on "mileage since the last failure", " component age, "car age" and "car mileage." These features are then added to the datasets.

\subsection{Data-driven model and evaluation}
In the proposed data-driven model, we aim to predict whether or not a railcar component will need replacement after the designated cut-off time. To achieve this goal, we build a data-driven model that corresponds to each component. The input for this model includes the engineered features of the railcars before the cut-off date. The random forest model is selected to predict the need for a component replacement after the cut-off date. We chose the random forest as the classifier because it can incorporate categorical and quantitative features, which is the case in our proposed framework. 

To examine the effect of including PCA features, we compare the AUC of B-PCA-K datasets (dataset with PCA features while keeping" car age ", “car mileage ", “component age" and" mileage since the last failure") with the AUC of the W-PCA dataset (dataset without PCA features). We then evaluate the AUC of the B-PCA-NK dataset (the dataset with the PCA feature, without keeping" car age", " car mileage", " component age", and" mileage since the last failure"). We also analyze the over-sampling method using Adaptive Synthetic (ADASYN) without PCA and ADASYN with PCA (ADASYN-PCA). The ADASYN method is used to handle imbalanced datasets. The fundamental concept of ADASYN is to utilize a weighted distribution for different minority class samples based on their level of difficulty in learning \cite{yang2021improved,he2008adasyn}. The results are shown in Table~\ref{tab:auc-result}. 

\begin{table}[h]
\centering
\caption{10-fold AUC of different models by components.} 
\label{tab:auc-result}
\begin{tabular}{ccccc}
\hline
Model & Component \#1 & Component \#2 & Component \#3 & Component \#4 \\
\hline
Without-PCA & 0.65 & 0.67 & 0.58 & 0.77\\
B-PCA-K & 0.67& 0.69 & 0.61 & 0.79\\
B-PCA-NK & 0.61 & 0.65 & 0.59 & 0.69\\
ADASYN & 0.63 & 0.70 & 0.57 & 0.80\\
ADASYN-PCA & 0.63 & 0.70 & 0.57 & 0.79\\
\hline
\end{tabular}
\end{table}

\begin{figure*}[hbt!]
    \centering
    \begin{subfigure}[b]{0.475\textwidth}
        \centering
        \includegraphics[width=\textwidth]{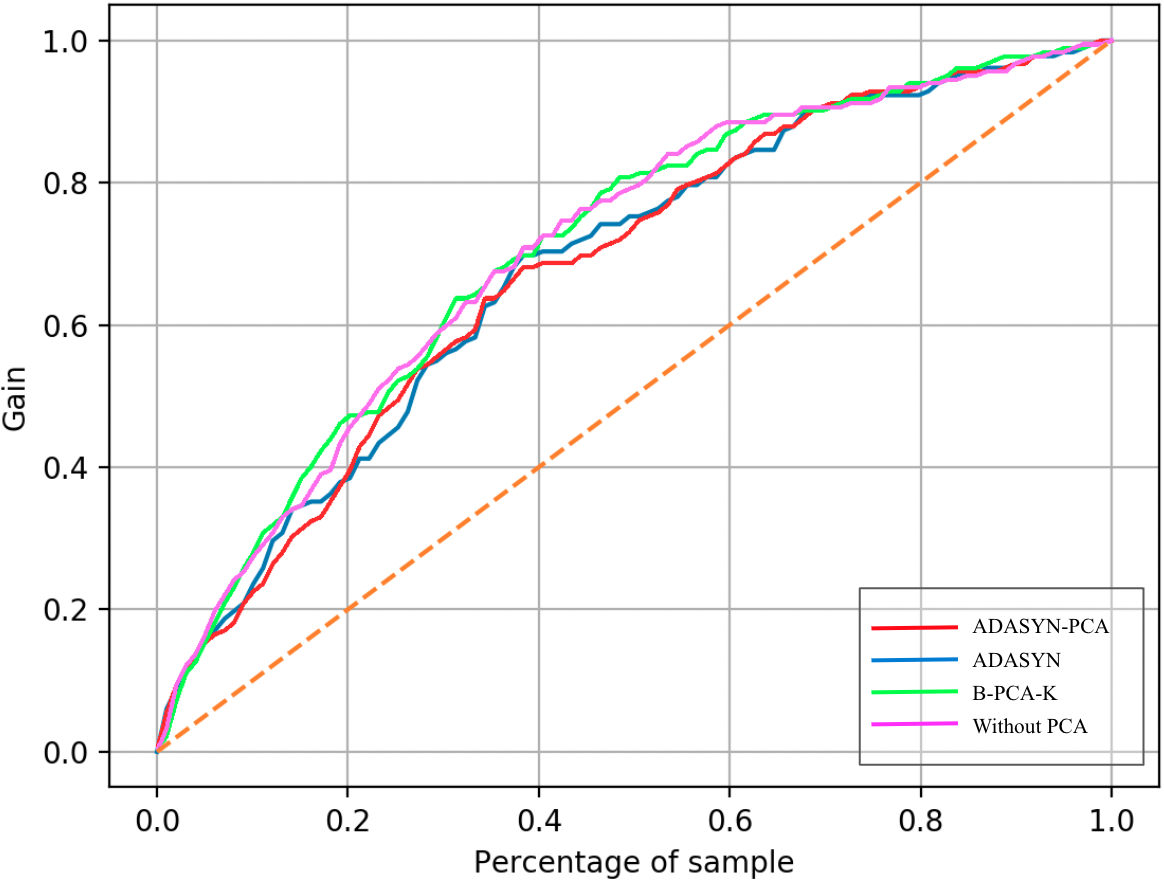}
        \caption[]%
        {{\small Component \#1}}    
        \label{fig:gain-c1}
    \end{subfigure}
    \hfill
    \begin{subfigure}[b]{0.475\textwidth}  
        \centering 
        \includegraphics[width=\textwidth]{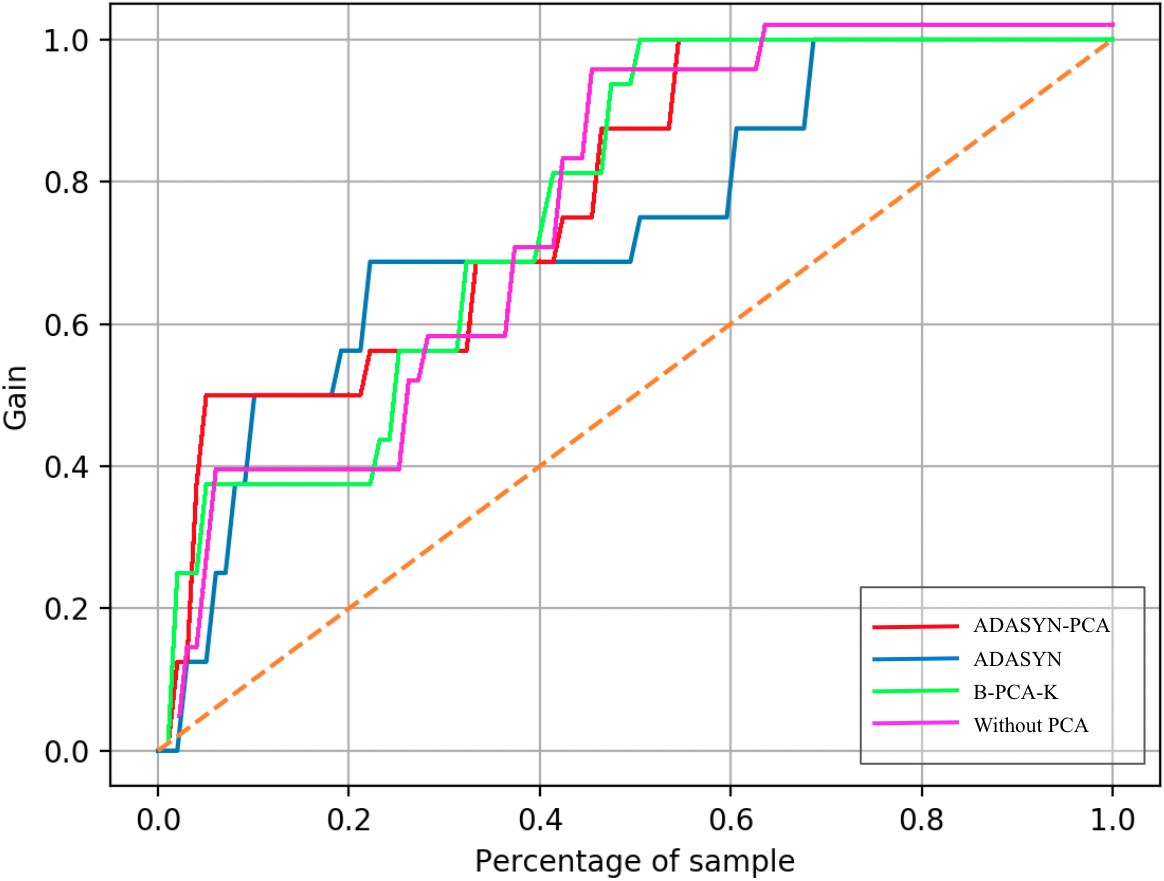}
        \caption[]%
        {{\small Component \#2}}    
        \label{fig:gain-c2}
    \end{subfigure}
    \vskip\baselineskip
    \begin{subfigure}[b]{0.475\textwidth}   
        \centering 
        \includegraphics[width=\textwidth]{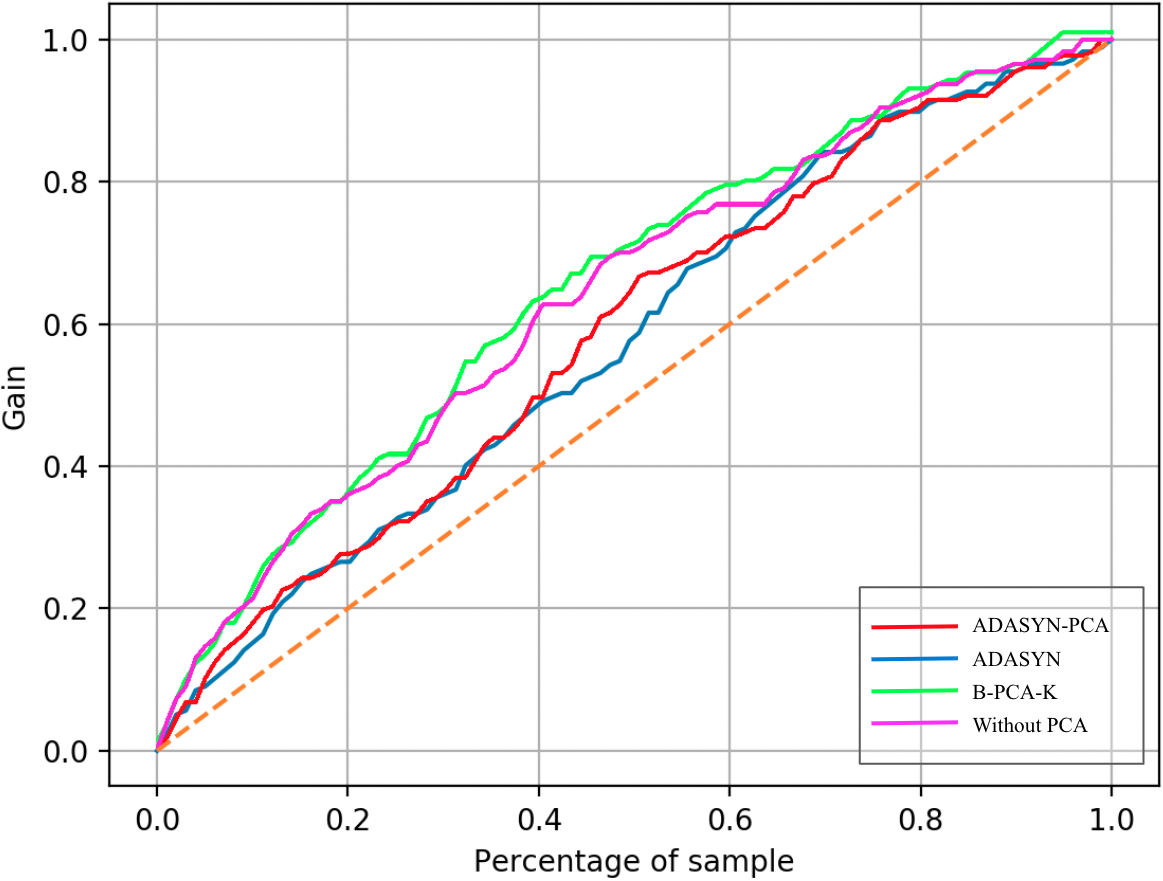}
        \caption[]%
        {{\small Component \#3}}    
        \label{fig:gain-c3}
    \end{subfigure}
    \hfill
    \begin{subfigure}[b]{0.475\textwidth}   
        \centering 
        \includegraphics[width=\textwidth]{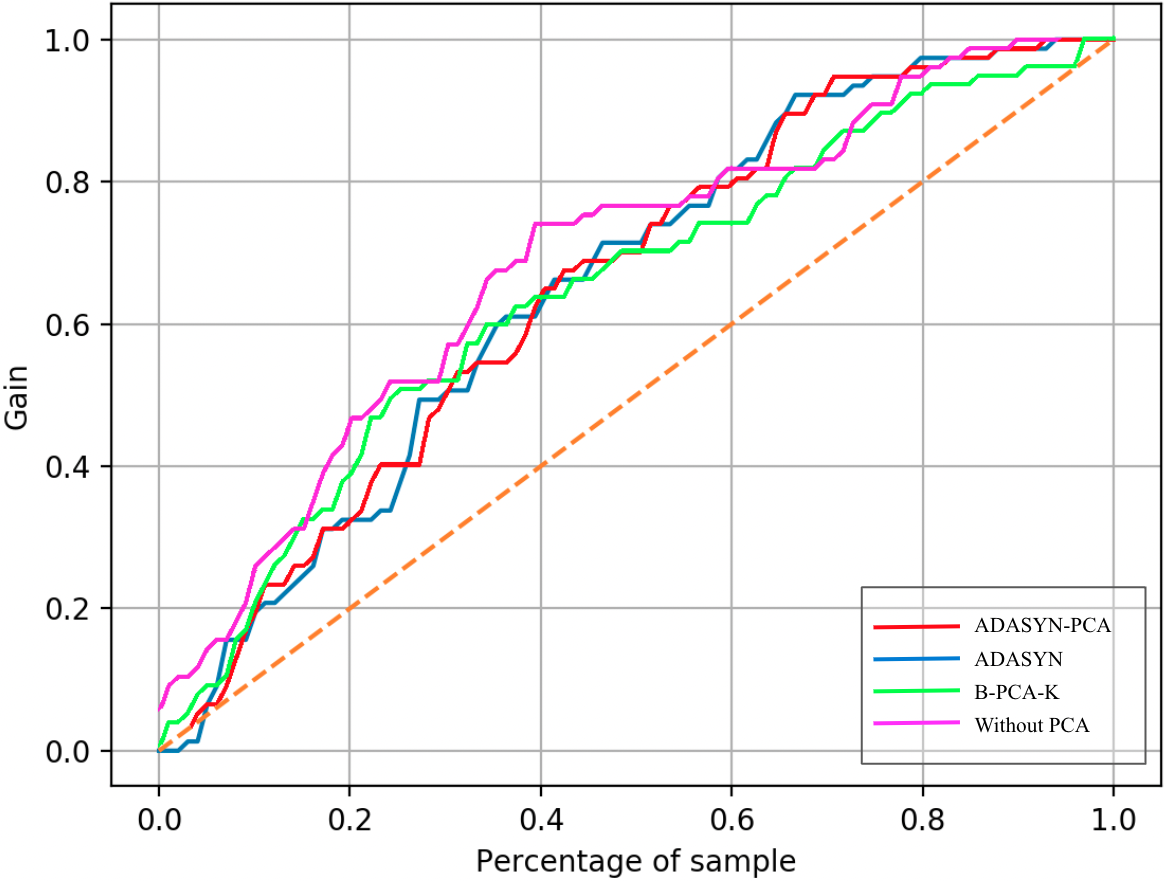}
        \caption[]%
        {{\small Component \#4}}    
        \label{fig:gain-c4}
    \end{subfigure}
    \caption[]
    {\small Gain chart for different component groups} 
    \label{fig:gain-chart}
\end{figure*}

Looking at the results in Table~\ref{tab:auc-result}, we notice that, in every case, the AUC of B-PCA-K is greater than the datasets without PCA features. Also, from the values of B-PCA-NK, we conclude we need to keep "car age", "car mileage", "component age", and "mileage since last failure" in the dataset. It is observed that using PCA and oversampling (ADASYN) simultaneously improves the value of AUC. As an additional comparison metric, Gain Chart is used to study the data-driven models with different settings, such as Without-PCA, B-PCA-K, and ADASYN-PCA, as shown in Figure~\ref{fig:gain-chart}. 

\begin{figure*}[h]
    \centering
    \begin{subfigure}[b]{0.475\textwidth}   
        \centering 
        \includegraphics[width=\textwidth]{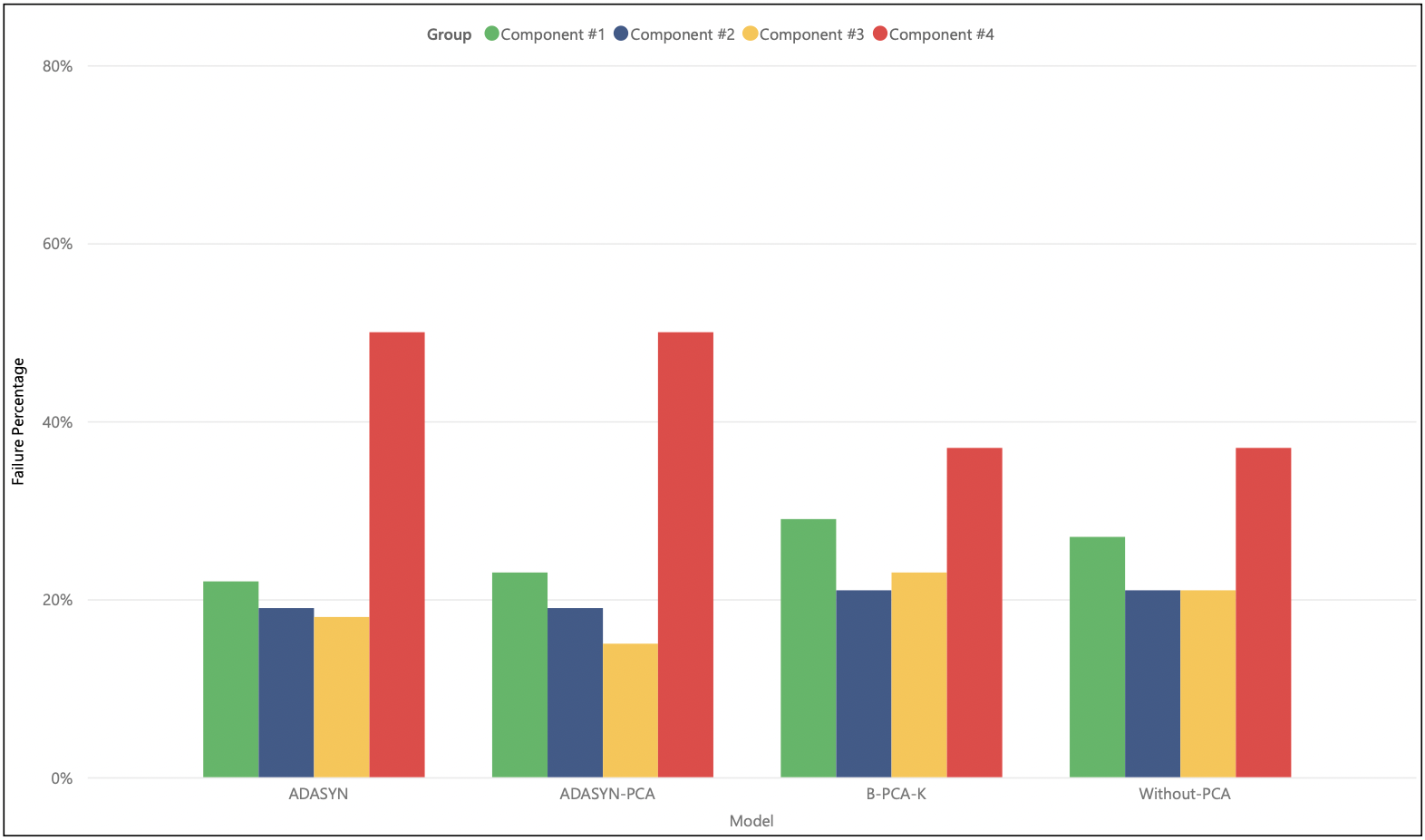}
        \caption[]%
        {{\small Prediction accuracy}}    
        \label{fig:fr-1}
    \end{subfigure}
    \hfill
    \begin{subfigure}[b]{0.475\textwidth}   
        \centering 
        \includegraphics[width=\textwidth]{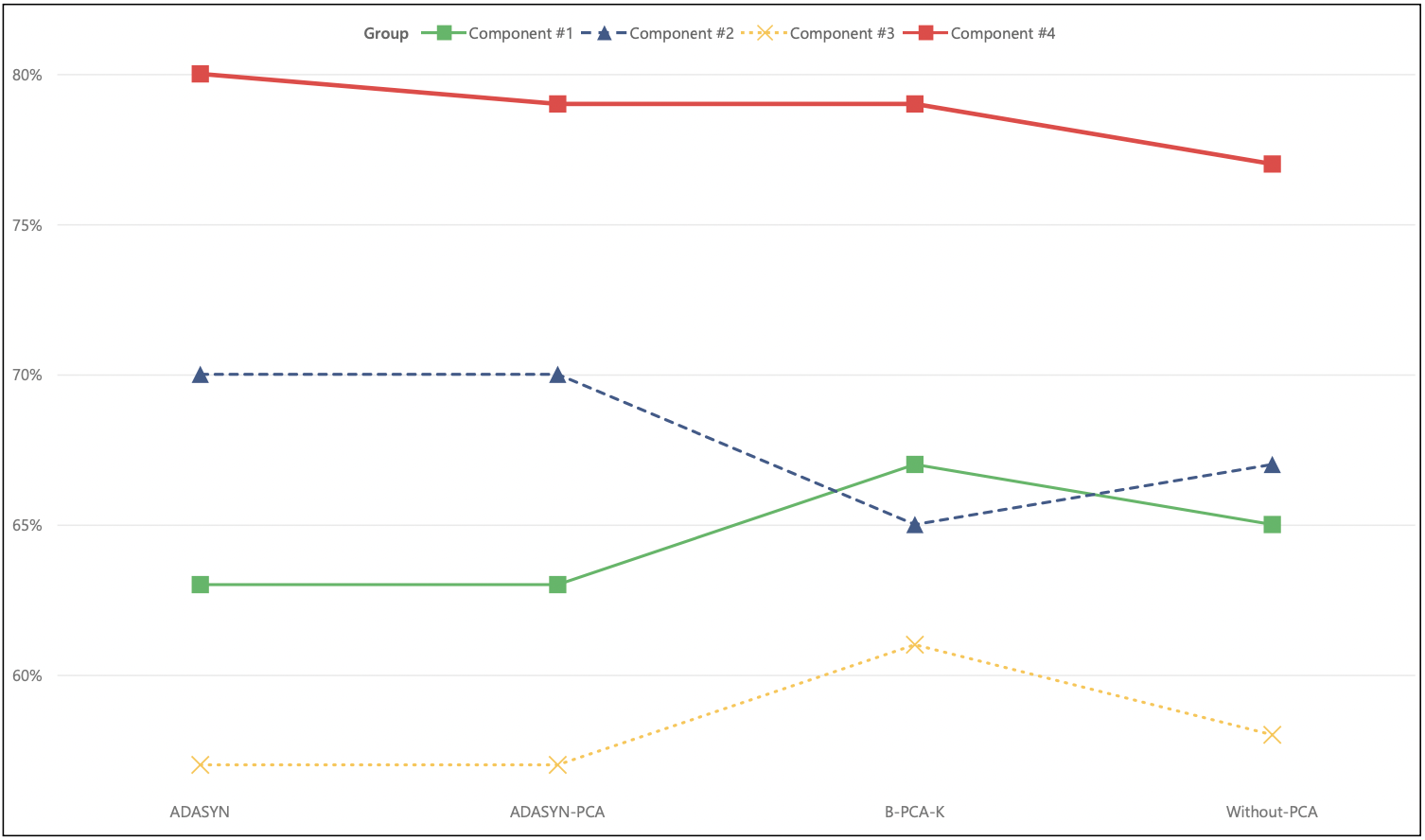}
        \caption[]%
        {{\small AUC metrics}}    
        \label{fig:auc-2}
    \end{subfigure}
    \caption[]
    {\small Failure percentage and AUC metrics which can be captured by top 10\% of the sample.} 
  \label{fig:mix_results}
\end{figure*}

These figures compare the number of failures captured by the selected samples to the random selection model. For instance, Figure~\ref{fig:mix_results} shows the number of failures detected when NARO selects 10 percent of the samples, or, as shown in Table  \ref{tab:auc-result}, with the combination of PCA and ADASYN, we can detect 63\%  of failure within 10 percent of samples in the component \#1 dataset.

\section{Conclusion and future research} \label{sec:Conclusion}

This paper develops a machine learning-based predictive maintenance expert system to assign a health rating to each railcar within a specific fleet for a NARO in the United States. This proposed predictive maintenance model contains three segments: feature, data-driven, and health score models. In the feature model, we utilized the data associated with four components of a railcar as NARO experts deem them to be the most critical components for this model.  We then collected railcar maintenance history and component data from 1984 to 2020.  This dataset includes almost sixty features. The dataset, in its raw form, was appropriate for preventive maintenance methods. However, in the predictive maintenance methods, which is the focus of our study, historical data was utilized to predict whether or not a component in a railcar would need replacement within the next 12 months. Therefore, to generate the required dataset for our proposed framework, we split historical data into four different datasets, representing the data corresponding to the components that we investigated. To predict future failures of these components, January 2019 was defined as the cut-off date. We then created and extracted new features based on records before that cut-off date. The dependent variable (target) was defined by whether or not a railcar component failed after January 2019. In the feature model, "component age", "car age", "mileage since last replacement", and "car mileage" were consistently in the top five essential features across all datasets. In the next step, we used PCA to create two PCA features based on "mileage since the last failure", "component age", "car age" and "car mileage" to add to the datasets.

The data-driven expert model predicts whether or not a railcar component will need replacement after the designated cut-off date. We chose a random forest classifier, given that categorical and quantitative features existed in our model. To measure the effectiveness of adding PCA features, we compared our model with PCA features to four different models, including the model without PCA, ADASYN, ADASYN with PCA, and model with PCA features by removing original features.

We compared these models using two criteria, namely, AUC and Gain Chart. Finally, in the health score model, we calculated each railcar's health score based on the outcome of each component's data-driven model. Our expert model was able to detect 96.4\% within 50\% of the sample. This indicates that our method is effective in diagnosing failures in a fleet of railcars. The health score was then defined as the weighted average of the failure probability for the railcar components. By assigning health rates to each railcar, NARO can categorize the railcars based on their health rates and efficiently allocate their resources to inspect the railcars with high failure probabilities. An important direction for an extension of this research is to develop a cost optimization model within this framework.  The goal would be to provide a recommendation for specific assets to maintain based on their unique composite health rating, usage history, and long term marketability. The goal is to minimize the total cost of asset ownership while simultaneously ensuring the optimal availability of railcar assets. 


\bibliographystyle{elsarticle-num} 
\bibliography{cas-refs}

\end{document}